\pdfoutput=1

\documentclass[11pt]{article}

\usepackage[final]{acl}

\usepackage{times}
\usepackage{latexsym}
\usepackage{amsmath}
\usepackage{multirow}

\usepackage[T1]{fontenc}

\usepackage[utf8]{inputenc}

\usepackage{microtype}

\usepackage{inconsolata}

\usepackage{graphicx}

\title{Mars-PO: Multi-Agent Reasoning System Preference Optimization}

\author{Xiaoxuan Lou$^1$$^2$, Chaojie Wang$^1$\thanks{$^*$Corresponding author}, Bo An$^1$$^2$ \\
$^1$Skywork AI, $^2$Nanyang Technological University\\}

\begin{document}
\maketitle
\begin{abstract}
Mathematical reasoning is a fundamental capability for large language models (LLMs), yet achieving high performance in this domain remains a significant challenge. The auto-regressive generation process often makes LLMs susceptible to errors, hallucinations, and inconsistencies, particularly during multi-step reasoning. In this paper, we propose Mars-PO, a novel framework to improve the mathematical reasoning capabilities of LLMs through a multi-agent system. It combines high-quality outputs from multiple agents into a hybrid positive sample set and pairs them with agent-specific negative samples to construct robust preference pairs for training. By aligning agents with shared positive samples while addressing individual weaknesses, Mars-PO achieves substantial performance improvements on mathematical reasoning benchmarks. For example, it increases the accuracy on the MATH benchmark of the state-of-the-art instruction-tuned LLM, Llama3.1-8B-Instruct, from 50.38\% to 57.82\%. Experimental results further demonstrate that our method consistently outperforms other baselines, such as supervised fine-tuning, vanilla DPO, and its enhanced versions, highlighting the effectiveness of our approach. 
\end{abstract}

\begin{figure*}[t]
\centering
  \includegraphics[width=\textwidth]{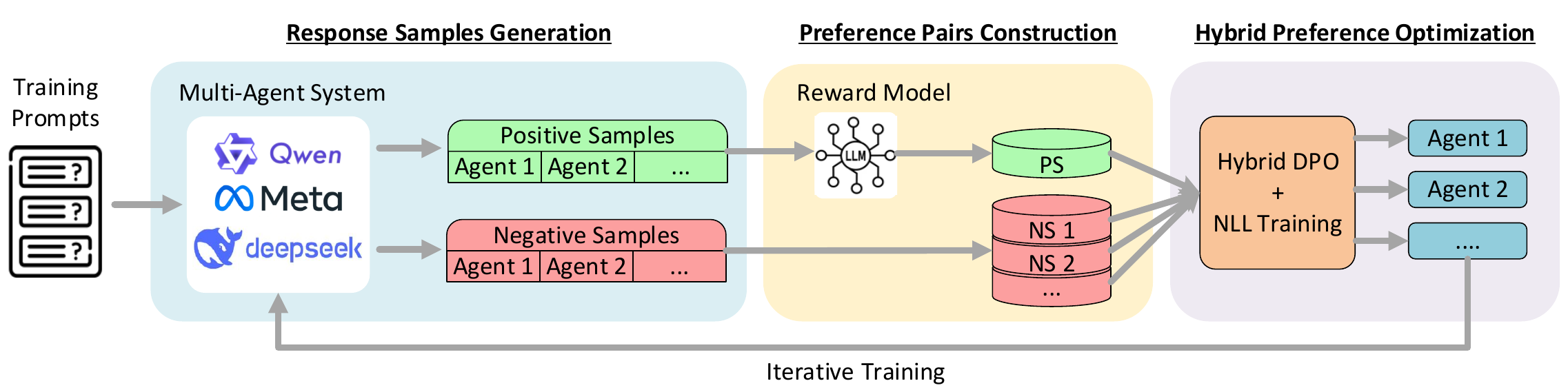}
  \caption{\textbf{Mars-PO Framework.} Our preference optimization method consists of three steps: (i) \emph{Response Samples Generation:} training prompts are fed into the multi-agent system to generate candidate responses, which are then classified as positive or negative for each agent based on answer correctness. (ii) \emph{Positive Pairs Construction:} positive samples from all agents are evaluated by a reward model to distill a high-quality positive sample set (PS) for the entire system, while negative samples (NS) proceed directly to the next step. (iii) \emph{Hybrid Preference Optimization:} preference pairs are selected to perform Mars-PO for each agent, supplemented by NLL loss and optional iterative training to improve model robustness and performance.}
  \label{fig:workflow}
\end{figure*}

\section{Introduction}

\begin{figure*}[t]
  \includegraphics[width=0.48\linewidth]{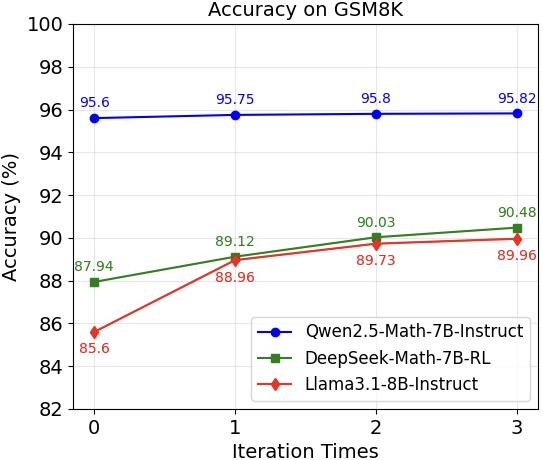} \hfill
  \includegraphics[width=0.48\linewidth]{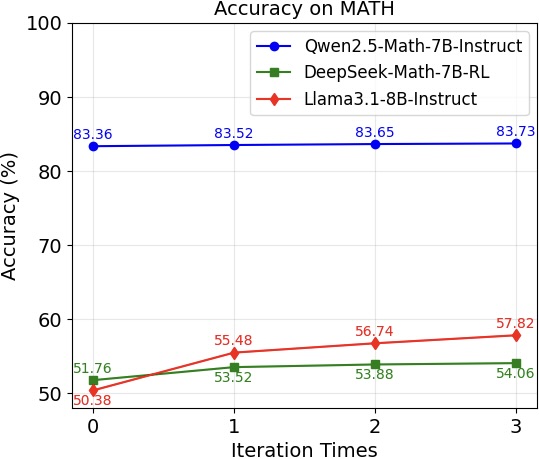}
  \caption {Accuracy of iterative Mars-PO training on GSM8K and Math.}\label{fig:first}
\end{figure*}

Mathematical reasoning is a critical yet highly challenging task for large language models (LLMs) \cite{yu2023metamath,lu2024mathgenie,luo2023wizardmath,wang2023mathcoder,shao2024deepseekmath,lu2024step,lai2024step}. It requires not only strong foundational knowledge in mathematics but also the ability to perform precise computations \cite{yu2023metamath,touvron2023llama}, logical reasoning \cite{lu2024mathgenie,pang2024iterative}, and multi-step problem-solving \cite{lu2024step,lai2024step}. Despite significant advancements in the capabilities of LLMs, achieving robust and reliable mathematical reasoning remains an open challenge. A primary factor contributing to this difficulty is the alignment of model-generated outputs with human preferences for correctness and clarity, particularly in complex domains like mathematical reasoning.

Among various alignment techniques, Direct Preference Optimization (DPO) \cite{rafailov2024direct} has emerged as a promising method for improving model behavior through preference-based training. It optimizes LLMs by leveraging preference signals derived from human or reward model judgments, directly adjusting the model's output distribution. DPO methods have demonstrated strong performance on general chat benchmarks, but their application to standard reasoning tasks often yields only moderate improvements or even performance degradation \cite{pang2024iterative,lu2024step,lai2024step}. Moreover, while DPO has achieved notable success in aligning single-agent systems, it frequently falls short in leveraging the collaborative potential of multi-agent systems. In such systems, diverse agents can contribute complementary strengths, which, if effectively utilized, could lead to the generation of higher-quality solutions.

To address these limitations, we propose a novel approach to achieve multi-agent reasoning system preference optimization, named as Mars-PO. This method extends the standard DPO framework to a multi-agent setting, leveraging the collective capabilities of multiple agents to improve alignment and reasoning performance. Figure \ref{fig:workflow} shows the framework of Mars-PO, which operates in three stages: 

\noindent\textbf{(i) Response Samples Generation:} Given a set of prompts, response samples are generated by multiple agents. These responses form the foundation for constructing positive and negative samples. By utilizing diverse agents, this process ensures that the generated responses capture a wide range of reasoning patterns and quality levels.

\noindent\textbf{(ii) Preference Pairs Construction:} Using the response samples, a reward model is employed to score all positive samples to extract a high-quality hybrid positive sample set. Then preference pairs are constructed by combining the hybrid positive sample set with agent-specific negative samples. This step is critical for encoding both shared strengths and individual weaknesses into the training process.
    
\noindent\textbf{(iii) Hybrid Preference Optimization:} Finally, the constructed preference pairs are used to train LLM agents via iterative preference optimization. By aligning the model with hybrid positive samples while addressing agent-specific weaknesses, this step ensures that the model achieves robust improvements in reasoning accuracy.

To evaluate the effectiveness of our method, we apply Mars-PO to a multi-agent system consisting of three instruction-tuned state-of-the-art mathematical LLMs: Qwen2.5-Math, DeepSeek-Math and Llama3.1. Extensive experiments on standard reasoning benchmarks, i.e., GSM8K and MATH, demonstrate that Mars-PO significantly improves the mathematical reasoning capabilities of LLM agents, outperforming baseline approaches like single-agent DPO and other advanced fine tuning methods, as shown in Figure \ref{fig:first}. Notably, our method can further push the state-of-the-art reasoning accuracy to new heights, with performance gains of up to about 8\%.
The results highlight the advantages of leveraging multi-agent collaboration to amplify performance gains and align models with task-specific requirements.
To sum up, our main contribution are as follows:
\begin{itemize}
    \item We propose a novel method, Mars-PO, that extends DPO to multi-agent systems for enhanced performance in mathematical reasoning tasks.
    \item We introduce a new strategy for constructing hybrid positive sample sets, combining the strengths of multiple agents to create a high-quality training dataset.
    \item Through rigorous evaluation, we demonstrate the effectiveness of Mars-PO in significantly improving the alignment and reasoning capabilities of LLM agents, setting a new benchmark in preference-based optimization for mathematical reasoning.
\end{itemize}

\section{Related Work}

\subsection{Mathematical Reasoning}

Large Language Models (LLMs) have demonstrated impressive reasoning abilities, driven by their auto-regressive nature, which enables them to accurately predict the next token using contextual information. However, these models still face significant challenges in handling sophisticated reasoning tasks, particularly in mathematical domains.

To address these limitations, prior research has explored various approaches to enhance the mathematical reasoning capabilities of LLMs. Several studies \cite{gao2023pal,chen2022program,zhou2023solving,yao2024tree} proposed advanced prompting methods based on the Chain-of-Thought (CoT) inference framework \cite{wei2022chain}, aiming to bring out LLMs’ mathematical skills without changing their parameters. 
In contrast, other methods aim to improve mathematical reasoning by optimizing LLM parameters through continued pretraining on large math-specific datasets \cite{azerbayev2023llemma,wang2023generative}, or fine-tuning with well constructed question-solution pairs \cite{yuan2023scaling,yue2023mammoth,wang2023mathcoder,luo2023wizardmath,gou2023tora,yang2023gpt,yu2023metamath,lu2024mathgenie}. These approaches significantly push LLMs to solve complex mathematical problems, achieving outstanding performance on benchmarks such as GSM8K \cite{cobbe2021training} and MATH \cite{hendrycks2021measuring}.

\subsection{Direct Preference Optimization}

While Reinforcement Learning from Human Feedback (RLHF) is widely used to align LLMs with human preference to improve prediction performance, the adopted reinforcement learning methods, like PPO \cite{schulman2017proximal}, pose a resource-intensive and time-consuming requirement on the reward model training. To address this issue, Direct Preference Optimization (DPO) \cite{rafailov2024direct} is proposed as a more efficient and equally effective alternative. DPO distinguishes itself
by enabling the model to learn a policy directly from user
preference data, eliminating the need for an explicit reward
function. 

While DPO has proven effective in chat benchmarks, it offers
only marginal benefits or may even negatively impact mathematical reasoning. Several previous works \cite{xu2024chatglm,jiao2024learning} uses DPO to improve model’s mathematical generation quality. Several previous works leveraged stepwise error annotations to further improve DPO's performance on mathematical reasoning tasks \cite{lu2024step,lai2024step}.

\section{Methodology}

Given a multi-agent system comprising multiple pretrained or instruction-tuned large language models, with access to a set of training inputs and the ability to evaluate the correctness of final outputs, our goal is to simultaneously enhance the performance of all agents within the system. To achieve the goal, we propose Hybrid Preference Optimization, which consists of three steps: (i) Response Samples Generation, (ii) Preference Pairs Construction and (iii) Hybrid Preference Optimization, as shown in Figure \ref{fig:workflow}. We provide additional details about the methodology design in the following.

\subsection{Response Samples Generation}

The generation of response samples lays the foundation for constructing high-quality preference pairs. In the context of a multi-agent system, this process involves coordinated sampling from multiple agents to ensure diverse and representative outputs for subsequent preference optimization.

We assume the target multi-agent system is composed of $K$ agents, denoted as $\{M_1, M_2, ..., M_K\}$. The used training dataset is denoted as $D=\{(x_i,y_i)\}$, where $x_i$ is the question prompt and $y_i$ is the corresponding correct response. Given the substantial performance improvements achieved by the CoT framework, mathematical reasoning tasks often utilize CoT reasoning steps $c_i$ to derive the final answer $a_i$. Hence, the target response $y_i$ can be expressed as a concatenation of $c_i$ and $a_i$, i.e., $y_i = (c_i, a_i)$. 
For each LLM agent $M_k$ ($1 \leq k \leq K$), we generate $N$ different responses for each input prompt $x_i \in D$, which are denoted as $y_i^n$: 
\begin{equation}
\label{eq:ans_eq}
  y_i^n = (c_i^n, a_i^n) \sim M_k(x_i)
\end{equation}
where $n \in \{1,2,...,N\}$. 

To evaluate the correctness of answers $a_i^n$ in the generated responses, we introduce a new boolean variable, $b_i^n$. Here, $b_i^n=1$ indicates that the $n$-th samples generated by agent $M_k$ contains a correct answer, i.e., $a_i^n=a_i$, while $b_i^n=0$ denotes an incorrect answer. Hence, based on the correctness of deduced answers, we can further divide those generated responses into two sets: 
\begin{equation}\label{eq:set_eq}
\begin{aligned}
    G_k^w = \{(c_i^n, a_i^n) | b_i^n=1\} \\
    G_k^l = \{(c_i^n, a_i^n) | b_i^n=0\}
\end{aligned}
\end{equation}
where $G_k^w$ denotes the set of positive (winning) samples with correct answers, while $G_k^l$ denotes the set of negative (losing) samples with incorrect answers.

\subsection{Preference Pairs Construction}

After identifying positive/negative sample sets of the multi-agent system, the next step is to construct preference pairs, a process that represents the most critical step in DPO-like methods. These pairs serve as the foundation for training reward-aligned language models. In our Mars-PO framework, we extend this process to incorporate multi-agent interactions, enabling the construction of a hybrid preference dataset that effectively leverages the complementary strengths of multiple agents. Specifically, we utilize a hybrid positive sample set combined with multiple agent-specific negative sample sets for the subsequent DPO training.  

The hybrid positive sample set $G^w$ is extracted from the outputs of all LLM agents, i.e., $G_k^w$ for $1 \leq k \leq K$. These agents generate candidate solutions for a shared set of mathematical reasoning tasks. A reward model, pre-trained to evaluate solution quality, assigns a reward score to each candidate. The highest-scoring outputs across all agents are merged into the hybrid positive sample set. This merging process ensures that the positive samples represent the best-performing solutions, irrespective of the agent of origin, thereby increasing the diversity and quality of the training data.

Unlike the hybrid positive set, these negative samples are agent-specific, reflecting each agent's unique failure modes. By pairing hybrid positive samples with negative samples tailored to individual agents, the constructed preference pairs expose the limitations of each agent while reinforcing the benefits of the shared positive solutions. This step is instrumental in aligning LLM agents to achieve superior performance on mathematical reasoning tasks, as validated by our experimental results.

\subsection{Hybrid Preference Optimization}

As the core component of our Mars-PO framework, hybrid preference optimization applies DPO method to each agent using a combination of a hybrid positive sample set $G^w$ and agent-specific negative sample sets $G_k^l$. This process leverages the strengths of multi-agent collaboration to enhance the reasoning capabilities of each individual agent. The loss function used for optimizing parameters of each agent is expressed as $\mathcal{L} = \mathcal{L}_{DPO} + \alpha \mathcal{L}_{NLL}$, where $\mathcal{L}_{DPO}$ and $\mathcal{L}_{NLL}$ can be expressed as:

\begin{equation}\label{eq:loss}
\begin{aligned}
  \mathcal{L}_{DPO} = -log \sigma(\beta log \frac{M_{\theta}(c_i^w, a_i^w | x_i)}{M_{k}(c_i^w, a_i^w | x_i)} \\
  - \beta log \frac{M_{\theta}(c_i^l, a_i^l | x_i)}{M_{k}(c_i^l, a_i^l | x_i)})
\end{aligned}
\end{equation}

\begin{equation}\label{eq:loss}
\begin{aligned}
  \mathcal{L}_{NLL} = \frac{log M_{\theta}(c_i^w, a_i^w | x_i)}{|c_i^w| +|y_i^w|}
\end{aligned}
\end{equation}
The Negative Log-likelihood Loss (NLL) is added to maintain base knowledge of the original agent model and prevent overfitting to preferences. To further enhance agent performance, we adopt an iterative training method to repeatedly update the parameters of the target LLMs.


\section{Experimental Setup}

In this section, we first present the evaluation LLM agents in the multi-agent system, which are also targets whose performance we aim to improve. Besides, we also introduce the used reward model for extracting high-quality positive samples and mathematical datasets used for reasoning tasks. Finally, we detail compared baselines used to highlight the advancement of our method. 

\subsection{Evaluation Models}

We evaluate the performance of Mars-PO on the multi-agent system consisting of three state-of-the-art instruction-tuned mathematical LLMs, including Qwen2.5-Math-Instruct \cite{yang2024qwen2} (with 7B parameters), Llama3.1-Instruct \cite{touvron2023llama} (with 8B parameters) and DeepSeek-Math-RL \cite{shao2024deepseekmath} (with 7B parameters). These three models have advanced mathematical reasoning capabilities to levels comparable to, or even surpassing, human performance. Among them, Qwen models stand out as significantly superior to their peers. Therefore, we also adopt the reward model used by Qwen, i.e., Qwen2.5-Math-RM-72B \cite{qwenreward}, as our reward model to score the quality of generated positive samples. 

\subsection{Reasoning Datasets}

Following previous research \cite{lu2024step,lai2024step,yu2023metamath,shao2024deepseekmath}, our evaluation performs on two mathematical reasoning datasets: GSM8K \cite{cobbe2021training} and MATH \cite{hendrycks2021measuring}. Both are classic benchmarks specifically designed to evaluate the arithmetic and word problem-solving capabilities of language models. They consist of challenging mathematical problems accompanied by well-structured reasoning steps leading to the correct answers.

\subsection{Compared Baselines}

We compare the performance of Mars-PO with four baselines: (i) original performance of the target instruction-tuned LLM agent, (ii) vanilla DPO method \cite{rafailov2024direct}, where each LLM agent uses their own preference pairs for post-training; (iii) DPO method combined with NLL item, which has been studied in previous works \cite{pang2024iterative} and can slightly improve reasoning capability of the target model; (iv) Supervised Fine Tuning (SFT) with extracted positive samples, which aims to investigate whether incorporating contrastive samples contributes to performance improvement.

\begin{table*}[t]
\centering
\resizebox{\linewidth}{!}{
\begin{tabular}{l|ll}
\hline
\hline
\multirow{2}{*}{LLM Agent} & \multicolumn{2}{c}{Benchmark Datasets} \\
\cline{2-3}
   & GSM8K & MATH \\
\hline
Qwen2.5-Math-7B-Instruct \hspace{4cm} & 95.60 \hspace{2cm}  & 83.36 \hspace{2cm}  \\
\hspace{1cm} + Mars-PO iter1 & $95.75^{\color{green}+0.15}$ & $83.52^{\color{green}+0.16}$ \\
\hspace{1cm} + Mars-PO iter2 & $95.79^{\color{green}+0.19}$ & $83.73^{\color{green}+0.37}$ \\
\hspace{1cm} + Mars-PO iter3 & $95.82^{\color{green}+0.22}$ & $83.65^{\color{green}+0.29}$ \\
\hspace{1cm} + Vanilla DPO & $89.61^{\color{red}-5.99}$ & $72.24^{\color{red}-11.12}$ \\
\hspace{1cm} + DPO+NLL & $90.14^{\color{red}-5.46}$ & $81.24^{\color{red}-2.12}$ \\
\hspace{1cm} + Postive SFT & $95.45^{\color{red}-0.15}$ & $83.22^{\color{red}-0.14}$ \\
\hline
DeepSeek-Math-7B-RL & 87.94 & 51.76 \\
\hspace{1cm} + Mars-PO iter1 & $89.12^{\color{green}+1.18}$ & $53.52^{\color{green}+1.76}$ \\
\hspace{1cm} + Mars-PO iter2 & $90.03^{\color{green}+3.09}$ & $53.88^{\color{green}+2.12}$ \\
\hspace{1cm} + Mars-PO iter3 & $90.48^{\color{green}+3.54}$ & $54.06^{\color{green}+3.30}$ \\
\hspace{1cm} + Vanilla DPO & $87.32^{\color{red}-0.61}$ & $51.44^{\color{red}-0.32}$ \\
\hspace{1cm} + DPO+NLL & $88.55^{\color{green}+0.61}$ & $51.52^{\color{red}-0.24}$ \\
\hspace{1cm} + Postive SFT & $88.17^{\color{green}+0.23}$ & $51.94^{\color{green}+0.18}$ \\
\hline
Llama3.1-8B-Instruct & 85.60 & 50.38 \\
\hspace{1cm} + Mars-PO iter1 & $88.96^{\color{green}+3.36}$ & $55.48^{\color{green}+5.10}$ \\
\hspace{1cm} + Mars-PO iter2 & $89.73^{\color{green}+4.13}$ & $56.74^{\color{green}+6.36}$ \\
\hspace{1cm} + Mars-PO iter3 & $89.96^{\color{green}+4.36}$ & $57.82^{\color{green}+7.44}$ \\
\hspace{1cm} + Vanilla DPO & $79.08^{\color{red}-6.52}$ & $42.48^{\color{red}-7.90}$ \\
\hspace{1cm} + DPO+NLL & $81.96^{\color{red}-3.64}$ & $43.08^{\color{red}-7.30}$ \\
\hspace{1cm} + Postive SFT & $86.50^{\color{red}-0.10}$ & $50.84^{\color{green}+0.46}$ \\
\hline
\hline
\end{tabular}}
\caption{Mathematical benchmark results of iterative Mars-PO using zero-shot greedy inference.}\label{tab:accuracy}
\end{table*}

\subsection{Implementation Details}

To sample responses from each agent, we follow previous works \cite{lu2024step,lai2024step,pang2024iterative,yu2023metamath} to use a zero-shot prompt that includes the question along with clear instructions to generate a chain-of-thought reasoning process. Ensure the response follows a specific format, making the final answer easy to identify and extract. We conduct three iterations of preference optimization to fully unlock the potential of LLM agents. In each iteration, we generate $N$ solutions ($N=40$ for GSM8K and $N=30$ for MATH) for each problem using sampling with temperature 0.8 for iterations 1 and temperature 1.2 for iterations 2-3, hoping for a substantial number of incorrect generations in the later iterations. 

The generated response samples are further processed to construct a hybrid positive sample set, extracted by the reward model, along with negative sample sets for each agent. Subsequently, we select 15 preference pairs from these sample sets for the following Mars-PO training.
The post-training of the target agent model is conducted over three epochs, with a batch size of 16 and a learning rate of 7e-7, using the AdamW optimizer.
The coefficient $\alpha$ and $\beta$ in Equation \ref{} are set as 1 and 0.1, respectively. Note that for iteration 2 and 3, $\beta$ is increased to 0.2 and 0.4, to further amplify the differences in the reward values of preference pairs.
All training is done using one node with eight A800 GPUs (80G memory).

\begin{figure*}[t]
  \includegraphics[width=0.48\linewidth]{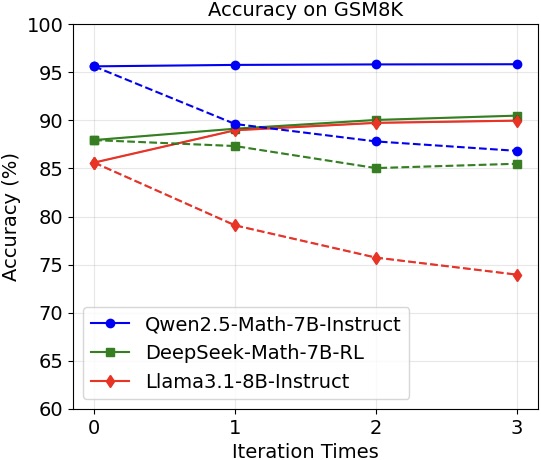} \hfill
  \includegraphics[width=0.48\linewidth]{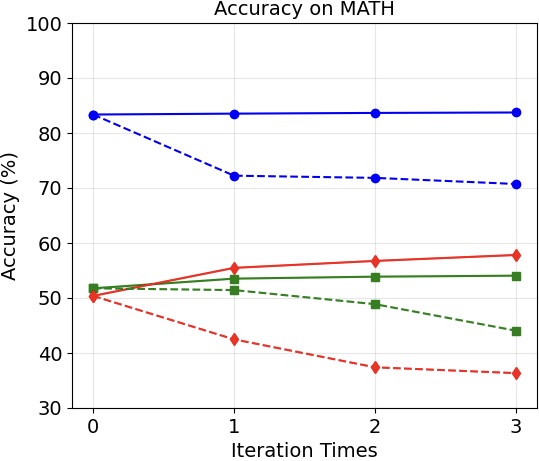}
  \caption {Accuracy comparison between vanilla DPO and Mars-PO. \textbf{Solid lines} represent results of Mars-PO method, while \textbf{dashed lines} represent results of traditional DPO method.}\label{fig:hybrid}
\end{figure*}

\section{Evaluation Results}

In this section, we first present the main results of Mars-PO in improving the performance of the multi-agent system, highlighting the advantages of our approach. We then compare it with various baselines to demonstrate the impact of the techniques incorporated into Mars-PO. Our experiments show that each of the introduced techniques contributes to a significant improvement in the performance of LLM agents.

\subsection{Main Results}

Table \ref{tab:accuracy} displays the prediction accuracy of LLM agents in the multi-agent system on GSM8K and MATH tasks. Note that these are the results after the first iteration of training. From the table, we can see the comparison between our method with four baselines. 
Experiment results demonstrate that Mars-PO consistently achieves higher accuracy across all baselines. Notably, our method even leads to a performance improvement of over 10\% on the Llama model, increasing the prediction accuracy on the challenging MATH dataset from 50.38\% to 55.48\%.

While conventional DPO results in a significant decline in model performance, even with the addition of a negative log-likelihood loss to partially alleviate the issue, it remains evident that vanilla DPO and its variants fail to further enhance the performance of state-of-the-art models. The reason behind this phenomenon could be that these models have already undergone extensive fine-tuning on widely available mathematical datasets, particularly GSM8K and MATH. As a result, the continued application of the DPO method results in severe overfitting. We also compare our approach with the SFT method, which directly utilizes the hybrid positive sample set, to demonstrate the necessity of contrastive optimization using agent-specific negative sample sets. Experimental results reveal that combining the hybrid positive sample set with agent-specific negative samples allows our method to generate more informative preference pairs, resulting in enhanced reasoning capabilities for LLM agents.

\subsection{Enhancement with iterative training}

To further improve model performance, we adopt an iterative training approach that progressively refines the model's reasoning capabilities. Iterative training involves multiple rounds of preference optimization, where each iteration builds upon the outputs and refinements from the previous round. This process allows the model to continually improve its understanding and alignment with high-quality reasoning patterns.

Table \ref{tab:accuracy} presents the prediction accuracy of the models across three iterations of training. As shown in the results, the accuracy generally increases with each iteration, demonstrating the effectiveness of iterative training in improving model performance. However, it is worth noting that in some cases, there is a slight drop in accuracy between certain iterations. Despite these minor fluctuations, the overall trend indicates a consistent upward trajectory in the model’s performance, highlighting the benefits of continued optimization through iterative training.


\subsection{Effect of hybrid positive samples}

To evaluate the impact of hybrid positive samples, we analyze their contribution to the overall model performance. Given hybrid positive samples are constructed by merging high-quality correct outputs from multiple agents, they can combine diverse strengths of all agents to create a unified and robust dataset. Hence, this approach is able to provide a richer and more comprehensive training signal compared to relying on positive samples from a single agent.

The comparison between the vanilla DPO method and our proposed Mars-PO demonstrates the effectiveness of hybrid positive samples in enhancing model performance, where Mars-PO consistently achieves higher accuracy than the vanilla DPO, as shown in in Table \ref{tab:accuracy}. Figure \ref{fig:hybrid} further illustrates the accuracy changes of the DPO and Mars-PO methods during the iterative training process. We can observe that Mars-PO consistently improves accuracy, while the vanilla DPO method results in a performance drop. These findings confirm that incorporating hybrid positive samples is a key factor in improving the performance of the model, making Mars-PO a more effective approach compared to traditional DPO.

\section{Conclusion}

In this paper, we proposed Hybrid Direct Preference Optimization (Mars-PO), a multi-agent framework to enhance the mathematical reasoning capabilities of large language models (LLMs). By combining a hybrid positive sample set with agent-specific negative samples, Mars-PO effectively leverages multi-agent collaboration to construct robust preference pairs for training. Experimental results demonstrate that this approach significantly improves LLM performance on mathematical reasoning benchmarks, showcasing the potential of hybrid preference optimization for complex reasoning tasks.

\bibliography{custom}

\end{document}